\documentclass[sigconf]{acmart} 
\usepackage{algpseudocode}
\usepackage{algorithm}
\usepackage{graphicx}
\usepackage{siunitx}
\usepackage{caption}
\usepackage{booktabs, multirow} 
\usepackage{tabularx}
\usepackage{csquotes}
\usepackage{wrapfig}
\usepackage{hyperref}
\usepackage{xcolor}

\AtBeginDocument{%
  \providecommand\BibTeX{{%
    \normalfont B\kern-0.5em{\scshape i\kern-0.25em b}\kern-0.8em\TeX}}}

\setcopyright{acmcopyright}
\copyrightyear{2023}
\acmYear{2023}
\acmDOI{10.1145/3583133.3590693}

\acmConference[GECCO '23]{the Genetic and Evolutionary Computation Conference 2023}{July 15--19, 2023}{Lisbon, Portugal}
\acmPrice{15.00}

\acmISBN{79-8-4007-0120-7/23/07}

\begin{document}

\title{Many-objective Optimization via Voting for Elites}


\author{Jackson Dean}
\orcid{0009-0007-3276-1890}
\email{jackson.dean@uvm.edu}
\affiliation{
  \institution{University of Vermont}
  \city{Burlington}
  \state{Vermont}
  \country{USA}
  \postcode{05401}
}
\author{Nick Cheney}
\orcid{0000-0002-7140-2213}
\email{ncheney@uvm.edu}
\affiliation{
  \institution{University of Vermont}
  \city{Burlington}
  \state{Vermont}
  \country{USA}
  \postcode{05401}
}

\renewcommand{\shortauthors}{Dean and Cheney}

\begin{abstract}


Real-world problems are often comprised of many objectives and require solutions that carefully trade-off between them.  Current approaches to many-objective optimization often require challenging assumptions, like knowledge of the importance/difficulty of objectives in a weighted-sum single-objective paradigm, or enormous populations to overcome the curse of dimensionality in multi-objective Pareto optimization.  Combining elements from Many-Objective Evolutionary Algorithms and Quality Diversity algorithms like MAP-Elites, we propose Many-objective Optimization via Voting for Elites (MOVE).  MOVE maintains a map of elites that perform well on different subsets of the objective functions.  On a 14-objective image-neuroevolution problem, we demonstrate that MOVE is viable with a population of as few as 50 elites and outperforms a naive single-objective baseline.  We find that the algorithm’s performance relies on solutions jumping across bins (for a parent to produce a child that is elite for a different subset of objectives).  We suggest that this type of goal-switching is an implicit method to automatic identification of stepping stones or curriculum learning.  We comment on the similarities and differences between MOVE and MAP-Elites, hoping to provide insight to aid in the understanding of that approach – and suggest future work that may inform this approach’s use for many-objective problems in general.

\end{abstract}

\keywords{Neuroevolution, Indirect Encoding, CPPN, Many-Objective Optimization, MAP-Elites, Stepping Stones}


\maketitle

\section{Introduction}
Complex real-world optimization problems often require solutions that carefully balance non-linear and non-intuitive trade-offs between many competing objectives.  As a result, evolutionary algorithms excel at finding solutions to multi-objective problems, with approaches such as Pareto optimization able to simultaneously explore a wide range of trade-offs between objectives and optimize each goal synergistically, in spite of, or agnostic to other competing objectives~\cite{deb2011multi}.


However evolving solutions to many-objective problems becomes increasingly challenging as the number of objectives grows \cite{coello2007evolutionary}. The population size required to maintain Pareto fronts scales exponentially and quickly becomes computationally infeasible~\cite{deb2005finding, ishibuchi2008effectiveness}.  Alternative approaches that collapse the many objectives down into a single weighted-sum objective or optimize them on a schedule may not properly weight/order each objective in the aggregate fitness function/schedule, as doing so would require knowledge of the dynamics and trade-offs between the objectives a priori~\cite{gomez1997incremental}.  
%
Existing approaches to optimize many-objectives (see~\cite{chand2015evolutionary} for a review) often assume that this privileged knowledge is available and assign different weights to the objectives. 

\textbf{Pareto dominance} is the traditional way to compare solutions on multiple objectives without weighting~\cite{deb2011multi}. Existing algorithms, such as NSGA-II \cite{deb2002fast} and SPEA2 \cite{zitzler2001spea2}, are effective at approximating a theoretical Pareto front (the set of non-dominated solutions) when there are few objectives. However, extrapolating from the population sizes determined in~\cite{pan2017region}, we estimate that we would need a population of over 8000 to optimize our 14 objectives via Pareto optimization. 
A previous attempt to improve on Pareto dominance for use with many objectives is PPD-MOEA which uses partial Pareto dominance on subsets of objectives~\cite{sato2010pareto}. Unlike our approach, PPD-MOEA uses two subsets, the ordering of objectives is fixed and predetermined, and is only tested up to 10 objectives.

Determining the attributes of goals that lead to intermediate \emph{stepping stones} towards a complex solution is a significant challenge for evolutionary algorithms
~\cite{lehman2011abandoning, woolley2011deleterious}.  
Open-ended novelty-seeking and Quality Diversity algorithms similarly seek to explore a diversity of novel solutions in hopes of discovering promising stepping stones, and may combine this with a pressure to become highly fit within a wide variety of behavioral/phenotype niches~\cite{pugh2016quality, lehman2011evolving, pugh2015confronting, cully2017quality, lehman2008exploiting, secretan2008picbreeder, stanley2017open, soros2014identifying, stanley2016strictness}.  

\textbf{MAP-Elites}~\cite{mouret2015illuminating} is popular \emph{quality diversity} algorithm with a map of discrete cells which each represent a phenotypic behavioral characteristic.  Each cell maintains a member of the population, the top performing (elite) individual on a global objective function that matches the subset of behavior characteristics assigned to that cell.  
This approach generates significant Quality Diversity, while implicitly overcoming the problem of determining stepping stones or learning curriculum order, as offspring for any given subset of the phenotypic space compete to become the elite in every other cell within the map -- constantly goal-switching to find potential stepping stones that may enable them to perform well on other phenotypic subsets~\cite{gaier2019quality, nguyen2015innovation}. 
%
%
{Multi-Objective MAP-Elites}~\cite{pierrot2022multi} fills each cell in the behavior map with a Pareto front rather than a single individual. 
Unlike in our work, each cell corresponds with a unique behavior descriptor and optimizes all objectives simultaneously.

\section{Approach}
Inspired by these approaches, we introduce a new variation of the MAP-Elites algorithm for optimizing many-objective search problems. Our approach, Many-objective Optimization via Voting for Elites (MOVE), defines each cell within a map as a subset of the many objectives that make up the overall search problem, and maintains a map of elites that are the best solutions found for each objective-combination in the map (Fig.~\ref{fig:trajectory}).  
While this approach inherits the structure, goal-switching, and diverse stepping-stones of the above approaches, its goal of finding the single optimal solution for the many-objective problem means that it is not exactly a quality diversity algorithm. Quality diversity algorithms seek to maximize the sum of fitness across all cells (phenotypic niches) in the map, while MOVE is more similar to approaches which construct building blocks that all ultimately lead to a single aggregate solution~\cite{forrest1993relative} as it explores many different subsets of the possible trade-offs and relationships between objectives in parallel. 
        
In MOVE, the population of genomes is stored in a map of cells. Each cell is assigned a fixed number of fitness functions selected randomly from all possible combinations. 
Every cell contains up to one elite, the most fit solution found thus far on the objectives in that cell. During selection and reproduction, each elite produces one child, a mutated copy of itself. 
For every cell on the map, each child is compared to the current elite 
by summing the number of fitness functions (of the cell’s assigned functions) on which the child scores higher than the elite. 
If the child earns more than half of these ‘votes’, it replaces the elite. 
Further intuition for MOVE is provided in Appendix~\ref{detailed_approach}
and a full implementation is available at \url{github.com/uvm-neurobotics-lab/MOVE}.
    
While this generic algorithm is agnostic to the particular type of solution being evolved, here we use a 14-dimensional optimization problem known to be fraught with local optima~\cite{woolley2011deleterious}: target image generation via Compositional Pattern Producing Networks~\cite{stanley2007compositional}.  
%
In the experiments below we set out to evolve synthetic images which maximize 14 different Image Quality Assessment metrics from the PyTorch Image Quality (PIQ) library~\cite{kastryulin2022piq} (Appendix~\ref{sec:obj_funcs}). 
    
    
    This approach allows evolution to exploit goal-switching, since a solution that performs well on one fitness function may produce offspring that perform well on different fitness functions. Unlike traditional scheduling approaches, no information about the ideal order of tasks is required as the algorithm designer does not have to choose when solutions jump between cells. 
    

    Results reported here employ the target image in Fig. \ref{tab:top_images}.
    Results from additional target images are reported in Appendix~\ref{other_targets}.
     Each condition includes 20 trials of 1000 generations. Unless otherwise noted, each trial used a map (i.e. population size) of 100 cells. We evaluate significance via a Wilcoxon rank-sum test with $\alpha=0.05$.

    To analyze how MOVE compares with a naive multi-objective approach, we conducted two baseline experiments:
    \textbf{Single-objective hillclimbers} each optimize only one of the 14 objectives. Each of the 14 hillclimbers produce 7 children per generation, approximating the computational costs 
    of MOVE.
        %
    The \textbf{all-objective hillclimber} control consist of a single hillclimber with 100 children per generation, resulting in the same total compute as MOVE. The fitness of a given image is the mean normalized fitness of all 14 objectives, as is the case in a traditional multi-term fitness function. 

\section{Results}

 \begin{figure}[!t]\centering
        \includegraphics[width=\linewidth]{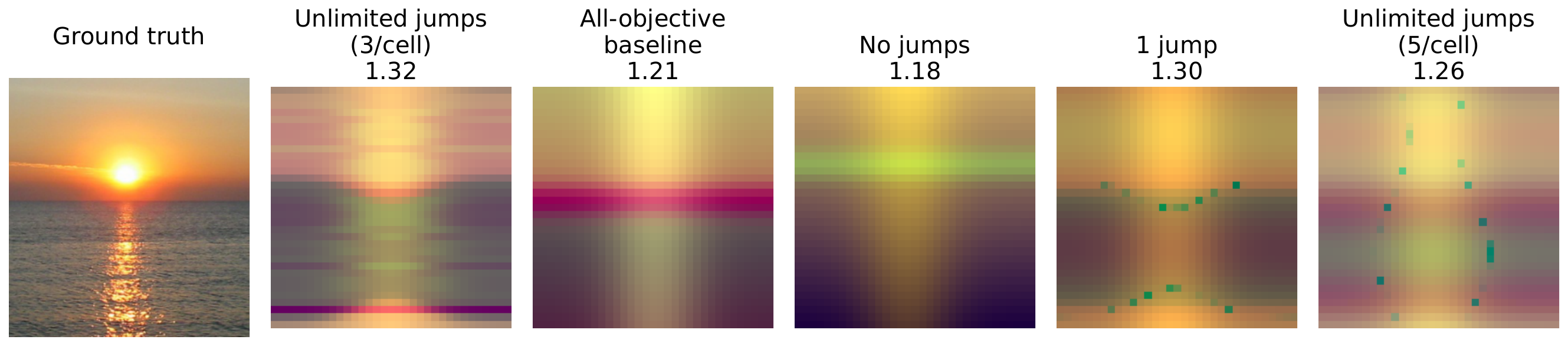}
        \captionof{figure}{Target image (left) and examples of evolved images under different experimental conditions.
        See Appendix~\ref{detailed_approach} for baselines results and additional target images.} 
        \label{tab:top_images}
        \Description{Images}
        \end{figure}

        \begin{table*}[!t]
        \centering
        \setlength\tabcolsep{6pt} 
        \begin{tabular}{| c | c | c | c | c | c | c |}
        \hline
        Functions & \ Total unique & Cells & Jumps  & Total & Jump & Champion
        \\
        per cell & solutions & in ancestry & in ancestry & replacements & proportion & fitness \\ 
        
            \hline
            1 & 12.75           & 6.73           & 24.57   & 261  & 47.43\%   & 1.10 \\ 
            3 & \textbf{40.95}  & 38.59        & 69.33   & 36,199  & 96.36\%   & \textbf{1.22} \\ 
            5 & 30.90           & \textbf{48.67} & \textbf{100.85} & \textbf{41,283}  & 97.40\%   & \textbf{1.19}  \\
            7 & 22.30           & 40.08 & \textbf{84.18}   & 28,891  & 97.78\% & \textbf{1.21} \\
            9 & 16.10           & 38.85 & \textbf{94.2}   & 28,092 & 98.30\%  & \textbf{1.20} \\
            11 & 11.75           & 32.27          & \textbf{101.36}   & 22,665  & \textbf{98.55\%} & 1.15 \\
        \hline

        \end{tabular}
        \captionof{table}{Statistics on the movement, diversity, and performance of solutions as a function of objectives per cell. The highest value (and those not significantly different from it) in each column are bold.  
        %
        %
        }
        \vspace{-0.5cm}
          \Description{Statistics about the movement, performance, and diversity of solutions.}
          \label{tab:numeric_results}
        \end{table*}

    \begin{figure*}[t]
      \centering
      \includegraphics[width=\linewidth]{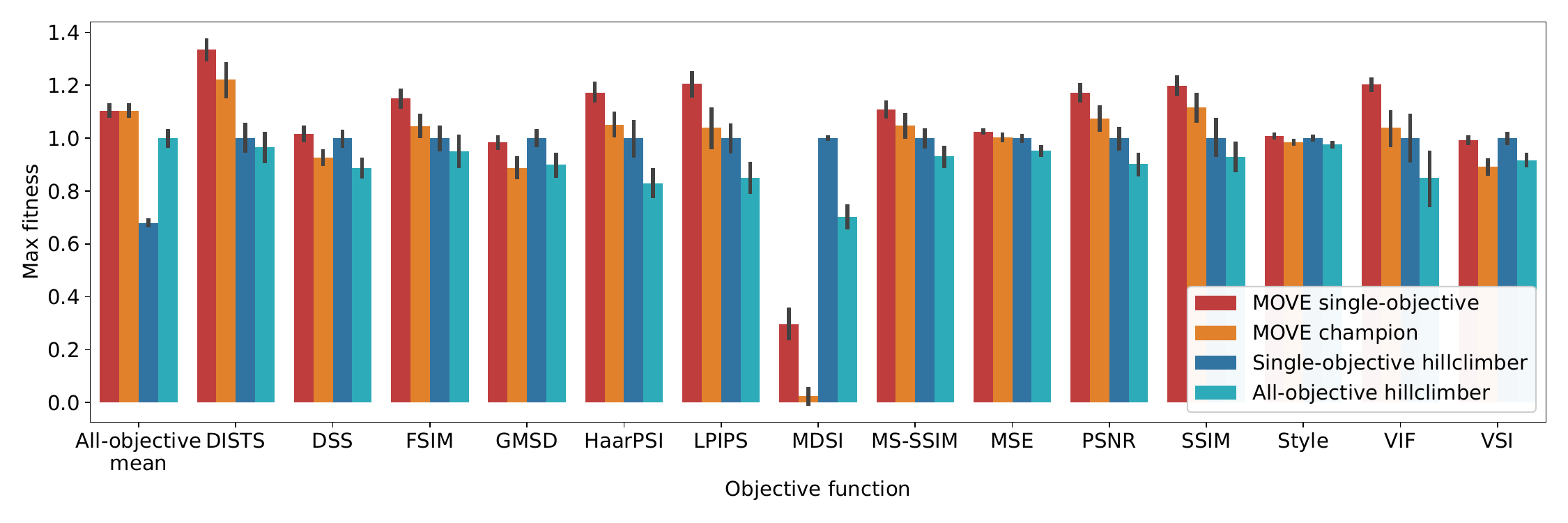}
      \vspace{-0.85cm}
      \caption{The most fit image across all objectives (left) and the most fit image on each individual objective function (remaining 14). All single-objective values are normalized relative to the single-objective hillclimber, while the all-objective result is normalized relative to the all-objective hillclimber. 
      %
      }
      \Description{Max fitness by objective function.}
        \label{fig:fn_max_fitness}
    \end{figure*}

    \subsection{Functions per cell} 
       While the general concept and flow of the MOVE algorithm is straightforward to describe, its ideal implementation and the impact of such decisions on performance relative to existing baselines are largely uncertain.  
       Firstly, it is not immediately clear how many objectives should be in each subset. Having more functions per cell would lead each subset to more closely approximate the full many-objective problem, while having fewer functions per cell would enable greater diversity across cells and potentially simpler subproblems to solve with fewer competing trade-offs in each.
       Here we explore subsets of 1, 3, 5, 7, 9, and 11 random objective functions per cell. Only odd numbers of functions were included to avoid ties. 
%
 
       With any number of functions per cell, MOVE  results in a significantly higher final overall fitness than the all-objective hillclimber (all $p < 0.01$; Table \ref{tab:numeric_results}). 
       The top performer according to overall fitness average across all 14 functions is the setting with 3 functions per cell, 
       but it was not significantly better than 5, 7, or 9.
        
        As the number of objective functions per cell grow, the number of objectives that any two cells have in common is expected to increase.  Thus one might expect it to become easier for an offspring to goal-switch with more functions per cell and perform well on a different cell than its parent.  In all cases, parents were far more likely to produce offspring that replaced other cells in the map than themselves (Table~\ref{tab:numeric_results}). 
        %
        If replacements happened randomly to each of the 100 cells, 99\% would be to a new (non-parent) cell. With 11 functions per cell, we see 98.55\% of replacements jumping cells.
        Even with just 3 functions per cell, we already see 96.26\% of cells replacing a parent other than their own, suggesting both the importance of goal switching above replacing parents -- and perhaps the potential of this algorithm to find many unique stepping stones and learning trajectories towards an ultimate solution. 

        Looking at the diversity within search, the total number of unique solutions across the 100 cells at the end of training was highly correlated with the number of functions per cell -- with 3 func./cell resulting in the greatest unique solutions ($p < 0.001$) and diversity of this final population dropping as func./cell increase. 

All trials in Table.~\ref{tab:numeric_results} employ 100 cells, but we also ran experiments with 25, 50 cells. With 5 functions per cell, MOVE worked just as well with 50 cells but worse with 25 (Appendix~\ref{sec:num_cells}), suggesting the potential even further computational savings than provided here. 
        
    \subsection{Goal-switching} 
        The results from the number of functions per cell experiments suggest that an important feature of MOVE is the ability of offspring to replace cells other than their parent cell. We hypothesized that this allows MOVE to find stepping stones and escape local optima along the path of any one cell in isolation. To investigate further, we conduct an ablation study that tests the importance of switching between cells in the map. 
        We explore three experimental conditions in which children were eligible to replace a different set of elites.
        
        \textbf{No jumping:} children can only replace their parents. This is functionally similar to 100 parallel hillclimbers. 
        \textbf{One jump:} children can replace any cell, including their parent, but only one. If a child is eligible to replace multiple elites, the one that lost the vote by the largest margin is chosen. 
        \textbf{Unlimited jumping:} children can replace any cell and replace multiple cells each generation.
        
        With 5 functions/cell, allowing unlimited jumping (all-objective fitness: 1.19) or a one jump (1.18) significantly increased performance compared with no jumping (1.09). 
        Allowing multiple jumps per offspring was the same (for 1, 7, and 9 functions per cell) or better (for 3, 5, and 11 functions per cell). These data support the hypothesis that MOVE benefits from the ability to goal-switch. 
%
%
        
        Furthermore, when an elite produced a surviving offspring, the replaced cell shared significantly more functions with the parent cell than would be expected from two random cells (Appendix~\ref{sec:goal_switching}), suggesting that shared objectives do enable increased goal-switching. Additionally, with 5 functions per cell, the path taken by the ancestors of a final solution included almost half of all the cells in the map on average, suggesting that a wide variety of cells serve as stepping stones en route to a final solution (Table~\ref{tab:numeric_results}, Appendix~\ref{detailed_results}).

    \subsection{By function}

        Table~\ref{tab:numeric_results} demonstrated that MOVE can achieve a higher mean performance across all objectives at once than an all-objective hillclimber attempting to optimize an aggregate function of all objectives.  But how does this translate to performance on the individual objectives?  In Fig.~\ref{fig:fn_max_fitness}, we compare the performance of MOVE (5 fn/cell; with unlimited jumping) to the all-objective hillclimber, and also to each single-objective hillclimber on how well each algorithm performs on each single objective.  
        MOVE significantly outperforms the aggregate hillclimber training on 13 of the 14 objectives.

        Surprisingly, MOVE also performs as well as, or better, than the single-objective hillclimber which does not have to compromise on any trade-offs between objectives on 9 of the 14 objective functions. This suggests that not only is MOVE able to overcome the trade-offs between objectives, it is able to find synergistic relationships between the different objectives and identify stepping stones to make the 14-dimensional many-objective optimization problem even easier than solving 14 independent single-objective searches. 
        
  

\section{Discussion}
    
    This paper offers a proof-of-concept for a novel way to optimize many-objectives, and tests the approach on an image neuroevolution problem. 
    Our method maintains solution diversity by separating a many-dimensional fitness landscape into parallel, but not fully independent, searches on smaller-dimensional spaces. Rather than maintaining a single many-dimensional Pareto front, MOVE stores a population of elites which each perform well on a different subset of the objective functions. This allows MOVE to successfully find stepping stones while maintaining a much smaller population than would be necessary in traditional Pareto optimization.

    This study finds important relationships between hyperparameters of this model and its overall performance.  
    The number of objectives per cell influenced overall aggregate fitness found, the propensity of offspring to replace their parents within the same cell vs goal-switching to become the elite in a new cell, and the diversity of solutions found (Table~\ref{tab:numeric_results}).

    Much like MAP-Elites~\cite{mouret2015illuminating} and Innovation Engines~\cite{nguyen2015innovation}, one of the driving forces behind the success of MOVE is the ability of solutions to jump between cells and generate their own trajectories of stepping stones over time (as seen in Appendix~\ref{fig:trajectory}).  We find that final solutions have used as many as 48\% of the cells in the map as stepping stones throughout their evolutionary trajectory -- switching goals as many as 100 times over the course of evolution (Table~\ref{tab:numeric_results}), and that removing the ability of MOVE to goal-switch across cells significantly hinders performance (Table ~\ref{tab:numeric_results}).  These migrations between cells evolving in parallel for different goals is also reminiscent of migration in island models~\cite{skolicki2005influence}. 
    The ease and order of jumps between cells may also inform weightings or orderings for other many-objective optimization approaches.  
    
    Future extensions of MOVE may incorporate concepts from multi-objective evolutionary algorithms, such as weighted objectives or other ways to condense multiple objectives into a single criterion to determine the ranking of an elite and a competing offspring in a cell.  Though, in our limited investigations, voting over objectives appeared to outperform normalized-weighted-sum fitness functions.
    Random objective weightings of many or all objectives (as opposed to the current binary inclusion of a random subset of functions) assigned to each cell in the map could be employed as a softened version of the voting mechanism.
    
    The results presented in this paper are likely highly dependent on the specific target images, encoding, and task chosen (see Appendix~\ref{other_targets} for experiments with other target images and~\ref{sec:future_work} for discussion of potential extensions).
    Future work should consider anti-correlated objectives, since MOVE may behave differently in problems where objectives are less aligned with one another than is the case with different Image Quality Assessment metrics here. 
    
    Despite its simplicity,
    MOVE is an effective method for many-objective optimization in this particular setting. 
    We are curious about the potential for this approach to generalize to the challenging problems of many-objective optimization across various domains.  




\begin{acks}
This material is based upon work supported by NSF Grant No. 2218063.  
Computations were performed on the Vermont Advanced Computing Core supported in part by NSF award No. 1827314. Thanks to Rachel Gehman for helpful brainstorming. 
\end{acks}

\bibliographystyle{ACM-Reference-Format}
\bibliography{main}


\begin{thebibliography}{29}


\ifx \showCODEN    \undefined \def \showCODEN     #1{\unskip}     \fi
\ifx \showDOI      \undefined \def \showDOI       #1{#1}\fi
\ifx \showISBNx    \undefined \def \showISBNx     #1{\unskip}     \fi
\ifx \showISBNxiii \undefined \def \showISBNxiii  #1{\unskip}     \fi
\ifx \showISSN     \undefined \def \showISSN      #1{\unskip}     \fi
\ifx \showLCCN     \undefined \def \showLCCN      #1{\unskip}     \fi
\ifx \shownote     \undefined \def \shownote      #1{#1}          \fi
\ifx \showarticletitle \undefined \def \showarticletitle #1{#1}   \fi
\ifx \showURL      \undefined \def \showURL       {\relax}        \fi
\providecommand\bibfield[2]{#2}
\providecommand\bibinfo[2]{#2}
\providecommand\natexlab[1]{#1}
\providecommand\showeprint[2][]{arXiv:#2}

\bibitem[\protect\citeauthoryear{Chand and Wagner}{Chand and Wagner}{2015}]%
        {chand2015evolutionary}
\bibfield{author}{\bibinfo{person}{Shelvin Chand} {and} \bibinfo{person}{Markus
  Wagner}.} \bibinfo{year}{2015}\natexlab{}.
\newblock \showarticletitle{Evolutionary many-objective optimization: A
  quick-start guide}.
\newblock \bibinfo{journal}{\emph{Surveys in Operations Research and Management
  Science}} \bibinfo{volume}{20}, \bibinfo{number}{2} (\bibinfo{year}{2015}),
  \bibinfo{pages}{35--42}.
\newblock


\bibitem[\protect\citeauthoryear{Coello, Lamont, Van~Veldhuizen,
  et~al\mbox{.}}{Coello et~al\mbox{.}}{2007}]%
        {coello2007evolutionary}
\bibfield{author}{\bibinfo{person}{Carlos A~Coello Coello},
  \bibinfo{person}{Gary~B Lamont}, \bibinfo{person}{David~A Van~Veldhuizen},
  {et~al\mbox{.}}} \bibinfo{year}{2007}\natexlab{}.
\newblock \bibinfo{booktitle}{\emph{Evolutionary algorithms for solving
  multi-objective problems}}. Vol.~\bibinfo{volume}{5}.
\newblock \bibinfo{publisher}{Springer}.
\newblock


\bibitem[\protect\citeauthoryear{Cully and Demiris}{Cully and Demiris}{2017}]%
        {cully2017quality}
\bibfield{author}{\bibinfo{person}{Antoine Cully} {and}
  \bibinfo{person}{Yiannis Demiris}.} \bibinfo{year}{2017}\natexlab{}.
\newblock \showarticletitle{Quality and diversity optimization: A unifying
  modular framework}.
\newblock \bibinfo{journal}{\emph{IEEE Transactions on Evolutionary
  Computation}} \bibinfo{volume}{22}, \bibinfo{number}{2}
  (\bibinfo{year}{2017}), \bibinfo{pages}{245--259}.
\newblock


\bibitem[\protect\citeauthoryear{Deb}{Deb}{2011}]%
        {deb2011multi}
\bibfield{author}{\bibinfo{person}{Kalyanmoy Deb}.}
  \bibinfo{year}{2011}\natexlab{}.
\newblock \bibinfo{booktitle}{\emph{Multi-objective optimisation using
  evolutionary algorithms: an introduction}}.
\newblock \bibinfo{publisher}{Springer}.
\newblock


\bibitem[\protect\citeauthoryear{Deb, Pratap, Agarwal, and Meyarivan}{Deb
  et~al\mbox{.}}{2002}]%
        {deb2002fast}
\bibfield{author}{\bibinfo{person}{Kalyanmoy Deb}, \bibinfo{person}{Amrit
  Pratap}, \bibinfo{person}{Sameer Agarwal}, {and} \bibinfo{person}{TAMT
  Meyarivan}.} \bibinfo{year}{2002}\natexlab{}.
\newblock \showarticletitle{A fast and elitist multiobjective genetic
  algorithm: NSGA-II}.
\newblock \bibinfo{journal}{\emph{IEEE transactions on evolutionary
  computation}} \bibinfo{volume}{6}, \bibinfo{number}{2}
  (\bibinfo{year}{2002}), \bibinfo{pages}{182--197}.
\newblock


\bibitem[\protect\citeauthoryear{Deb, Saxena, et~al\mbox{.}}{Deb
  et~al\mbox{.}}{2005}]%
        {deb2005finding}
\bibfield{author}{\bibinfo{person}{Kalyanmoy Deb}, \bibinfo{person}{Dhish~Kumar
  Saxena}, {et~al\mbox{.}}} \bibinfo{year}{2005}\natexlab{}.
\newblock \showarticletitle{On finding pareto-optimal solutions through
  dimensionality reduction for certain large-dimensional multi-objective
  optimization problems}.
\newblock \bibinfo{journal}{\emph{Kangal report}}  \bibinfo{volume}{2005011}
  (\bibinfo{year}{2005}), \bibinfo{pages}{1--19}.
\newblock


\bibitem[\protect\citeauthoryear{Forrest and Mitchell}{Forrest and
  Mitchell}{1993}]%
        {forrest1993relative}
\bibfield{author}{\bibinfo{person}{Stephanie Forrest} {and}
  \bibinfo{person}{Melanie Mitchell}.} \bibinfo{year}{1993}\natexlab{}.
\newblock \showarticletitle{Relative building-block fitness and the
  building-block hypothesis}.
\newblock In \bibinfo{booktitle}{\emph{Foundations of genetic algorithms}}.
  Vol.~\bibinfo{volume}{2}. \bibinfo{publisher}{Elsevier},
  \bibinfo{pages}{109--126}.
\newblock


\bibitem[\protect\citeauthoryear{Gaier, Asteroth, and Mouret}{Gaier
  et~al\mbox{.}}{2019}]%
        {gaier2019quality}
\bibfield{author}{\bibinfo{person}{Adam Gaier}, \bibinfo{person}{Alexander
  Asteroth}, {and} \bibinfo{person}{Jean-Baptiste Mouret}.}
  \bibinfo{year}{2019}\natexlab{}.
\newblock \showarticletitle{Are quality diversity algorithms better at
  generating stepping stones than objective-based search?}. In
  \bibinfo{booktitle}{\emph{Proceedings of the Genetic and Evolutionary
  Computation Conference Companion}}. \bibinfo{pages}{115--116}.
\newblock


\bibitem[\protect\citeauthoryear{Gomez and Miikkulainen}{Gomez and
  Miikkulainen}{1997}]%
        {gomez1997incremental}
\bibfield{author}{\bibinfo{person}{Faustino Gomez} {and} \bibinfo{person}{Risto
  Miikkulainen}.} \bibinfo{year}{1997}\natexlab{}.
\newblock \showarticletitle{Incremental evolution of complex general behavior}.
\newblock \bibinfo{journal}{\emph{Adaptive Behavior}} \bibinfo{volume}{5},
  \bibinfo{number}{3-4} (\bibinfo{year}{1997}), \bibinfo{pages}{317--342}.
\newblock


\bibitem[\protect\citeauthoryear{Ishibuchi, Tsukamoto, Hitotsuyanagi, and
  Nojima}{Ishibuchi et~al\mbox{.}}{2008}]%
        {ishibuchi2008effectiveness}
\bibfield{author}{\bibinfo{person}{Hisao Ishibuchi}, \bibinfo{person}{Noritaka
  Tsukamoto}, \bibinfo{person}{Yasuhiro Hitotsuyanagi}, {and}
  \bibinfo{person}{Yusuke Nojima}.} \bibinfo{year}{2008}\natexlab{}.
\newblock \showarticletitle{Effectiveness of scalability improvement attempts
  on the performance of NSGA-II for many-objective problems}. In
  \bibinfo{booktitle}{\emph{Proceedings of the 10th annual conference on
  Genetic and evolutionary computation}}. \bibinfo{pages}{649--656}.
\newblock


\bibitem[\protect\citeauthoryear{Kastryulin, Zakirov, Prokopenko, and
  Dylov}{Kastryulin et~al\mbox{.}}{2022}]%
        {kastryulin2022piq}
\bibfield{author}{\bibinfo{person}{Sergey Kastryulin}, \bibinfo{person}{Jamil
  Zakirov}, \bibinfo{person}{Denis Prokopenko}, {and}
  \bibinfo{person}{Dmitry~V. Dylov}.} \bibinfo{year}{2022}\natexlab{}.
\newblock \bibinfo{title}{PyTorch Image Quality: Metrics for Image Quality
  Assessment}.
\newblock
\newblock
\urldef\tempurl%
\url{https://doi.org/10.48550/ARXIV.2208.14818}
\showDOI{\tempurl}


\bibitem[\protect\citeauthoryear{Lehman and Stanley}{Lehman and
  Stanley}{2011a}]%
        {lehman2011abandoning}
\bibfield{author}{\bibinfo{person}{Joel Lehman} {and} \bibinfo{person}{Ken
  Stanley}.} \bibinfo{year}{2011}\natexlab{a}.
\newblock \showarticletitle{Abandoning objectives: Evolution through the search
  for novelty alone}.
\newblock \bibinfo{journal}{\emph{Evolutionary computation}}
  \bibinfo{volume}{19}, \bibinfo{number}{2} (\bibinfo{year}{2011}),
  \bibinfo{pages}{189--223}.
\newblock


\bibitem[\protect\citeauthoryear{Lehman and Stanley}{Lehman and
  Stanley}{2011b}]%
        {lehman2011evolving}
\bibfield{author}{\bibinfo{person}{Joel Lehman} {and}
  \bibinfo{person}{Kenneth~O Stanley}.} \bibinfo{year}{2011}\natexlab{b}.
\newblock \showarticletitle{Evolving a diversity of virtual creatures through
  novelty search and local competition}. In
  \bibinfo{booktitle}{\emph{Proceedings of the 13th annual conference on
  Genetic and evolutionary computation}}. \bibinfo{pages}{211--218}.
\newblock


\bibitem[\protect\citeauthoryear{Lehman, Stanley, et~al\mbox{.}}{Lehman
  et~al\mbox{.}}{2008}]%
        {lehman2008exploiting}
\bibfield{author}{\bibinfo{person}{Joel Lehman}, \bibinfo{person}{Kenneth~O
  Stanley}, {et~al\mbox{.}}} \bibinfo{year}{2008}\natexlab{}.
\newblock \showarticletitle{Exploiting open-endedness to solve problems through
  the search for novelty.}. In \bibinfo{booktitle}{\emph{ALIFE}}.
  \bibinfo{pages}{329--336}.
\newblock


\bibitem[\protect\citeauthoryear{Mouret and Clune}{Mouret and Clune}{2015}]%
        {mouret2015illuminating}
\bibfield{author}{\bibinfo{person}{Jean-Baptiste Mouret} {and}
  \bibinfo{person}{Jeff Clune}.} \bibinfo{year}{2015}\natexlab{}.
\newblock \showarticletitle{Illuminating search spaces by mapping elites}.
\newblock \bibinfo{journal}{\emph{arXiv preprint arXiv:1504.04909}}
  (\bibinfo{year}{2015}).
\newblock


\bibitem[\protect\citeauthoryear{Nguyen, Yosinski, and Clune}{Nguyen
  et~al\mbox{.}}{2015}]%
        {nguyen2015innovation}
\bibfield{author}{\bibinfo{person}{Anh Nguyen}, \bibinfo{person}{Jason
  Yosinski}, {and} \bibinfo{person}{Jeff Clune}.}
  \bibinfo{year}{2015}\natexlab{}.
\newblock \showarticletitle{Innovation engines: Automated creativity and
  improved stochastic optimization via deep learning}. In
  \bibinfo{booktitle}{\emph{Proceedings of the 2015 Annual Conference on
  Genetic and Evolutionary Computation}}. \bibinfo{pages}{959--966}.
\newblock


\bibitem[\protect\citeauthoryear{Pan, He, Tian, Su, and Zhang}{Pan
  et~al\mbox{.}}{2017}]%
        {pan2017region}
\bibfield{author}{\bibinfo{person}{Linqiang Pan}, \bibinfo{person}{Cheng He},
  \bibinfo{person}{Ye Tian}, \bibinfo{person}{Yansen Su}, {and}
  \bibinfo{person}{Xingyi Zhang}.} \bibinfo{year}{2017}\natexlab{}.
\newblock \showarticletitle{A region division based diversity maintaining
  approach for many-objective optimization}.
\newblock \bibinfo{journal}{\emph{Integrated Computer-Aided Engineering}}
  \bibinfo{volume}{24}, \bibinfo{number}{3} (\bibinfo{year}{2017}),
  \bibinfo{pages}{279--296}.
\newblock


\bibitem[\protect\citeauthoryear{Pierrot, Richard, Beguir, and Cully}{Pierrot
  et~al\mbox{.}}{2022}]%
        {pierrot2022multi}
\bibfield{author}{\bibinfo{person}{Thomas Pierrot}, \bibinfo{person}{Guillaume
  Richard}, \bibinfo{person}{Karim Beguir}, {and} \bibinfo{person}{Antoine
  Cully}.} \bibinfo{year}{2022}\natexlab{}.
\newblock \showarticletitle{Multi-objective quality diversity optimization}. In
  \bibinfo{booktitle}{\emph{Proceedings of the Genetic and Evolutionary
  Computation Conference}}. \bibinfo{pages}{139--147}.
\newblock


\bibitem[\protect\citeauthoryear{Pugh, Soros, and Stanley}{Pugh
  et~al\mbox{.}}{2016}]%
        {pugh2016quality}
\bibfield{author}{\bibinfo{person}{Justin~K Pugh}, \bibinfo{person}{Lisa
  Soros}, {and} \bibinfo{person}{Kenneth~O Stanley}.}
  \bibinfo{year}{2016}\natexlab{}.
\newblock \showarticletitle{Quality diversity: A new frontier for evolutionary
  computation}.
\newblock \bibinfo{journal}{\emph{Frontiers in Robotics and AI}}
  (\bibinfo{year}{2016}), \bibinfo{pages}{40}.
\newblock


\bibitem[\protect\citeauthoryear{Pugh, Soros, Szerlip, and Stanley}{Pugh
  et~al\mbox{.}}{2015}]%
        {pugh2015confronting}
\bibfield{author}{\bibinfo{person}{Justin~K Pugh}, \bibinfo{person}{Lisa~B
  Soros}, \bibinfo{person}{Paul~A Szerlip}, {and} \bibinfo{person}{Kenneth~O
  Stanley}.} \bibinfo{year}{2015}\natexlab{}.
\newblock \showarticletitle{Confronting the challenge of quality diversity}. In
  \bibinfo{booktitle}{\emph{Proceedings of the 2015 Annual Conference on
  Genetic and Evolutionary Computation}}. \bibinfo{pages}{967--974}.
\newblock


\bibitem[\protect\citeauthoryear{Sato, Aguirre, and Tanaka}{Sato
  et~al\mbox{.}}{2010}]%
        {sato2010pareto}
\bibfield{author}{\bibinfo{person}{Hiroyuki Sato},
  \bibinfo{person}{Hern{\'a}n~E Aguirre}, {and} \bibinfo{person}{Kiyoshi
  Tanaka}.} \bibinfo{year}{2010}\natexlab{}.
\newblock \showarticletitle{Pareto partial dominance MOEA and hybrid archiving
  strategy included CDAS in many-objective optimization}. In
  \bibinfo{booktitle}{\emph{IEEE Congress on Evolutionary Computation}}. IEEE,
  \bibinfo{pages}{1--8}.
\newblock


\bibitem[\protect\citeauthoryear{Secretan, Beato, D~Ambrosio, Rodriguez,
  Campbell, and Stanley}{Secretan et~al\mbox{.}}{2008}]%
        {secretan2008picbreeder}
\bibfield{author}{\bibinfo{person}{Jimmy Secretan}, \bibinfo{person}{Nicholas
  Beato}, \bibinfo{person}{David~B D~Ambrosio}, \bibinfo{person}{Adelein
  Rodriguez}, \bibinfo{person}{Adam Campbell}, {and} \bibinfo{person}{Kenneth~O
  Stanley}.} \bibinfo{year}{2008}\natexlab{}.
\newblock \showarticletitle{Picbreeder: evolving pictures collaboratively
  online}. In \bibinfo{booktitle}{\emph{Proceedings of the SIGCHI conference on
  human factors in computing systems}}. \bibinfo{pages}{1759--1768}.
\newblock


\bibitem[\protect\citeauthoryear{Skolicki and De~Jong}{Skolicki and
  De~Jong}{2005}]%
        {skolicki2005influence}
\bibfield{author}{\bibinfo{person}{Zbigniew Skolicki} {and}
  \bibinfo{person}{Kenneth De~Jong}.} \bibinfo{year}{2005}\natexlab{}.
\newblock \showarticletitle{The influence of migration sizes and intervals on
  island models}. In \bibinfo{booktitle}{\emph{Proceedings of the 7th annual
  conference on Genetic and evolutionary computation}}.
  \bibinfo{pages}{1295--1302}.
\newblock


\bibitem[\protect\citeauthoryear{Soros and Stanley}{Soros and Stanley}{2014}]%
        {soros2014identifying}
\bibfield{author}{\bibinfo{person}{L Soros} {and} \bibinfo{person}{Kenneth
  Stanley}.} \bibinfo{year}{2014}\natexlab{}.
\newblock \showarticletitle{Identifying necessary conditions for open-ended
  evolution through the artificial life world of chromaria}. In
  \bibinfo{booktitle}{\emph{ALIFE 14: The Fourteenth International Conference
  on the Synthesis and Simulation of Living Systems}}. MIT Press,
  \bibinfo{pages}{793--800}.
\newblock


\bibitem[\protect\citeauthoryear{Stanley}{Stanley}{2007}]%
        {stanley2007compositional}
\bibfield{author}{\bibinfo{person}{Kenneth~O Stanley}.}
  \bibinfo{year}{2007}\natexlab{}.
\newblock \showarticletitle{Compositional pattern producing networks: A novel
  abstraction of development}.
\newblock \bibinfo{journal}{\emph{Genetic programming and evolvable machines}}
  \bibinfo{volume}{8}, \bibinfo{number}{2} (\bibinfo{year}{2007}),
  \bibinfo{pages}{131--162}.
\newblock


\bibitem[\protect\citeauthoryear{Stanley, Cheney, and Soros}{Stanley
  et~al\mbox{.}}{2016}]%
        {stanley2016strictness}
\bibfield{author}{\bibinfo{person}{Kenneth~O Stanley}, \bibinfo{person}{Nick
  Cheney}, {and} \bibinfo{person}{Lisa~B Soros}.}
  \bibinfo{year}{2016}\natexlab{}.
\newblock \showarticletitle{How the strictness of the minimal criterion impacts
  open-ended evolution}. In \bibinfo{booktitle}{\emph{Artificial Life
  Conference Proceedings}}. MIT Press, \bibinfo{pages}{208--215}.
\newblock


\bibitem[\protect\citeauthoryear{Stanley, Lehman, and Soros}{Stanley
  et~al\mbox{.}}{2017}]%
        {stanley2017open}
\bibfield{author}{\bibinfo{person}{Kenneth~O Stanley}, \bibinfo{person}{Joel
  Lehman}, {and} \bibinfo{person}{Lisa Soros}.}
  \bibinfo{year}{2017}\natexlab{}.
\newblock \showarticletitle{Open-endedness: The last grand challenge you’ve
  never heard of}.
\newblock \bibinfo{journal}{\emph{While open-endedness could be a force for
  discovering intelligence, it could also be a component of AI itself}}
  (\bibinfo{year}{2017}).
\newblock


\bibitem[\protect\citeauthoryear{Woolley and Stanley}{Woolley and
  Stanley}{2011}]%
        {woolley2011deleterious}
\bibfield{author}{\bibinfo{person}{Brian~G Woolley} {and}
  \bibinfo{person}{Kenneth~O Stanley}.} \bibinfo{year}{2011}\natexlab{}.
\newblock \showarticletitle{On the deleterious effects of a priori objectives
  on evolution and representation}. In \bibinfo{booktitle}{\emph{Proceedings of
  the 13th annual conference on Genetic and evolutionary computation}}.
  \bibinfo{pages}{957--964}.
\newblock


\bibitem[\protect\citeauthoryear{Zitzler, Laumanns, and Thiele}{Zitzler
  et~al\mbox{.}}{2001}]%
        {zitzler2001spea2}
\bibfield{author}{\bibinfo{person}{Eckart Zitzler}, \bibinfo{person}{Marco
  Laumanns}, {and} \bibinfo{person}{Lothar Thiele}.}
  \bibinfo{year}{2001}\natexlab{}.
\newblock \showarticletitle{SPEA2: Improving the strength Pareto evolutionary
  algorithm}.
\newblock \bibinfo{journal}{\emph{TIK-report}}  \bibinfo{volume}{103}
  (\bibinfo{year}{2001}).
\newblock


\end{thebibliography}


\clearpage

\appendix

\section*{Appendix:}

\renewcommand{\thefigure}{A\arabic{figure}}
\renewcommand{\thetable}{A\arabic{table}}

\begin{figure}[t!]
          \includegraphics[width=\linewidth]{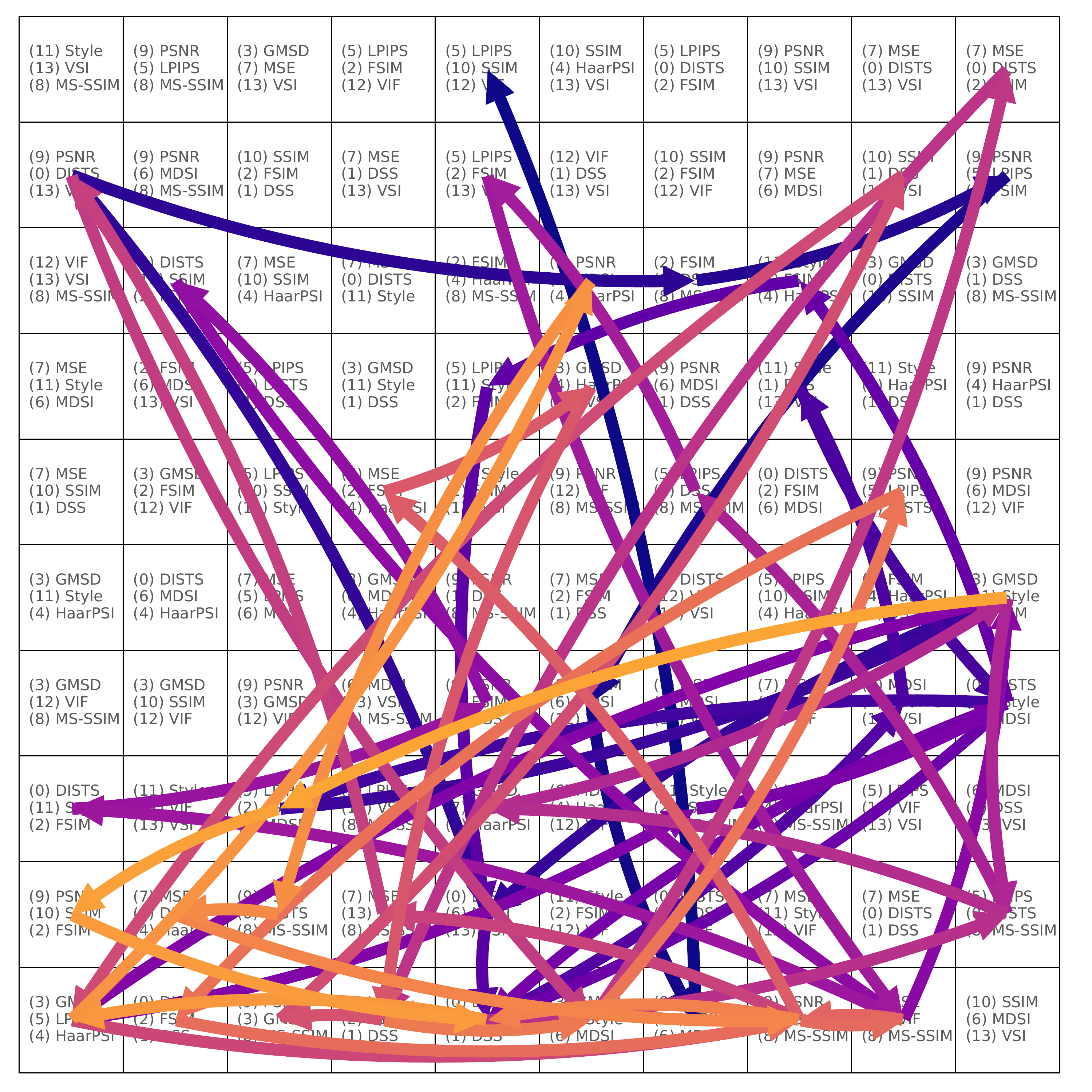}
          \caption{The evolutionary lineage of a solution to a 14-objective problem found by the MOVE algorithm.  Each bin in the map optimizes for a different subset of the overall objectives, and contains the single best solution found thus far (the elite) on that subset of objectives.  The arrows indicate the jumps (63 in total) between 40 unique cells in the lineage leading up to the champion of this run (colored warm to cool over evolutionary time). 
          We find that this map of elites is both efficient (able to use a far smaller population size than Pareto optimization) and effective (outperforming a single-objective hillclimber that optimizes for the sum of all 14 fitness functions together) and in 9/14 cases even matching or outperforming a hillclimber optimizing for only that individual fitness function).  
          The goal switching observed here is found to be a critical part of the optimization process, as performance drops significantly when elites are restricted to replacing existing solutions only within their own cell.
          }
          \Description{}
            \label{fig:trajectory}
\end{figure}


\section{Detailed Approach}
\label{detailed_approach}
\subsection{Algorithm}

\algnewcommand{\IIf}[1]{\State\algorithmicif\ #1\ \algorithmicthen}
\algnewcommand{\EndIIf}{\unskip\ \algorithmicend\ \algorithmicif}
    \begin{algorithm}
    \begin{algorithmic}[1]
        \caption{Many-objective Optimization via Voting for Elites (MOVE)}\label{euclid}
        \Require objectives $\gets$ a set of many fitness functions to optimize
        \Require $n \gets \mbox{number of objective functions per cell}$
        \Require $m \gets \mbox{number of cells in map}$
        \For{$i$ \texttt{in} $1,\ldots, m$}
            \State objective\_subsets$_i$$ \gets$ rand\_choice(objectives, $size=n$)
        \EndFor
        \State elites $\gets$ initial population of size m
        \For{generation \texttt{in} total generations}
            \State children $\gets []$
            \For{parent \texttt{in} elites}
                \State children.append(mutate(parent))
            \EndFor
            \For{child \texttt{in} children}
                \For {$i$ \texttt{in} $1,\ldots,m$}
                    \State votes $\gets 0$
                    \For{metric \texttt{in} objective\_subsets$_i$}
                        \State $F \gets$ fitness(elites$_i$, metric) 
                        \State $F^\prime \gets$ fitness(child, metric) 
                        \IIf {$F^\prime>\mbox{F}$} votes $\gets$ votes + 1
                    \EndFor
                    \IIf {$votes >\frac{n}{2}$} elites$_i$ $\gets$ child
                \EndFor
            \EndFor
        \EndFor
        \label{alg:voting}
    \end{algorithmic}
    \end{algorithm}

Intuition for the MOVE algorithm is provided in Algorithm~\ref{alg:voting}, however these operations can be done much more efficiently in practice (see \url{github.com/uvm-neurobotics-lab/MOVE} for an open-source implementation).

\subsection{Encoding}
Similar to~\cite{nguyen2015innovation}, we represent the individuals within our map as Compositional Pattern Producing Networks (CPPNs)~\cite{stanley2007compositional}.  
The CPPNs in these studies contained the activation functions: $sine$, $cosine$, $gaussian$, $identity$, and $sigmoid$. The inputs were $x$, $y$, and a $bias$ of $+1.0$. The outputs were $R, G, B$ color values \cite{stanley2007compositional}. No recombination/crossover was used.

\subsection{Objective functions}
\label{sec:obj_funcs}
The 14 target functions used were: Mean Squared Error (MSE), Peak Signal-to-Noise Ratio (PSNR), Feature Similarity Index (FSIM), Structural Similarity Index (SSIM), Multiscale Structural Similarity Index (MS-SSIM), Visual Information Fidelity (VIF), Visual Saliency-Induced index (VSI), Haar wavelet-based Perceptual Similarity Index (HaarPSI), Style loss, Gradient Magnitude Similarity Deviation (GMSD), Mean Deviation Similarity Index (MDSI), DCT Subband Similarity (DSS) Deep Image Structure and Texture Similarity (DISTS), and Learned Perceptual Image Patch Similarity (LPIPS). See~\cite{kastryulin2022piq} for a full description and references for all metrics. Metrics that measure image dissimilarity were inverted so that higher values were associated with more similar images. 

Many of the Image Quality Assessment metrics have been shown to strongly correlate with human perception. Some metrics focus on low-level structure like gradients and textures while others reward for high-level similarities. Newer approaches leverage deep neural network feature extractors trained for other tasks and compare images based on differences in extracted features. 

Fitness values were normalized within each target and function by the mean highest fitness found by the all-objective hillclimber for that function and target. Normalizing the range of each fitness function enables more intuitive visualization and analysis of individual-objective and mean-overall fitness values (e.g. Figure~\ref{fig:fn_max_fitness}) though the algorithm does not rely on these normalized values to assess the overall quality across multiple objectives, as majority voting across individual objectives in a cell is employed during fitness evaluation/selection.  An alternative version of this algorithm that explored the alternative method of determining fitness by the mean-normalized-fitness of objectives within a cell was also tested, and resulted in similar performance -- so the simpler and scale-agnostic voting mechanism is used throughout.  

Figure~\ref{fig:top_image_by_fn} shows the top image found across 20 runs of each single-objective hillclimber for 4 different target images.

\subsection{Number of cells}
\label{sec:num_cells}

Results in this work are reported from trials with 100 cells in the MOVE map. However we also ran experiments with as few as 25 cells. With 5 functions per cell, the 25, 50, and 100 cell conditions all resulted in higher final overall fitness than the all-objective hillclimber ($p<0.001$). Runs with 25 cells ($1.15 \pm 0.03$) performed significantly worse ($p \leq 0.04$) than 50 ($1.19 \pm 0.03$) and 100 ($1.19 \pm 0.03$). Interestingly, 50 cells did not perform significantly worse than 100 cells ($p=0.83$), suggesting that MOVE can perform effectively with surprisingly small population sizes.  

\section{Detailed results}
\label{detailed_results}
\begin{figure*}[!htp]
    \centering
    \includegraphics[scale=0.20]{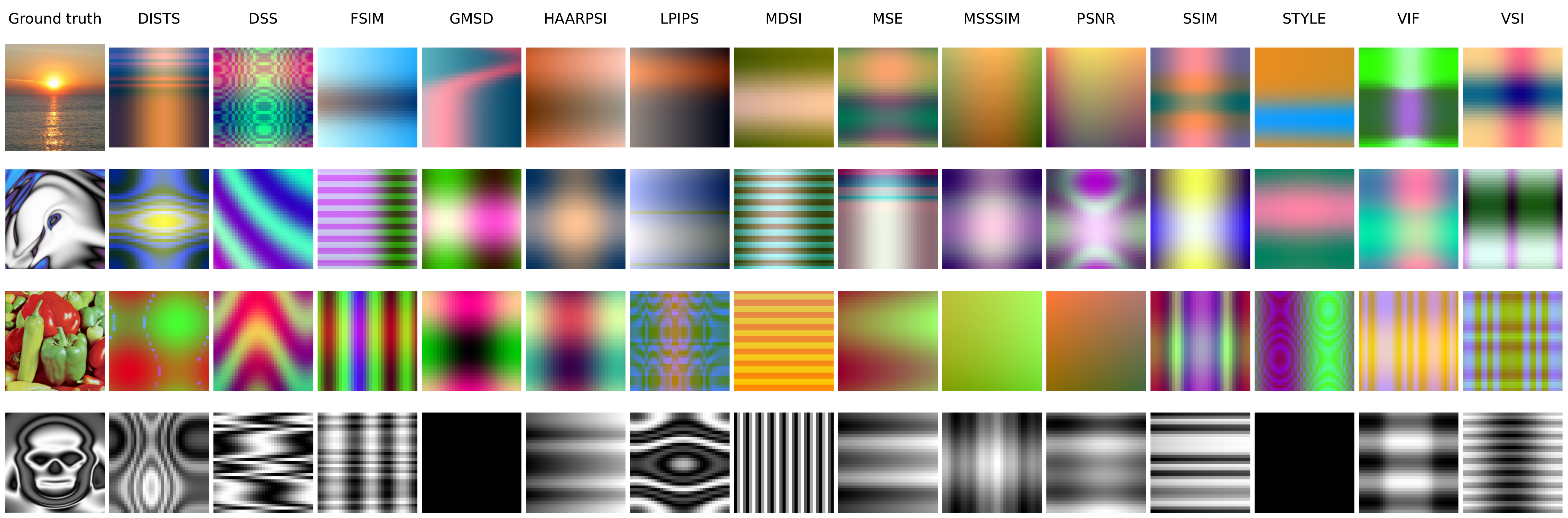}
    \caption{The top image found by any single-objective hillclimber for each objective. No single objective hillclimber creates a faithful reconstruction of the original due to the presence of deceiving local optima. However, by looking at the diversity of phenotypes within rows, it is apparent that each objective finds different local optima. Image of sunrise by \href{https://commons.wikimedia.org/wiki/User:Bogdan}{Bogdan} [\href{https://creativecommons.org/licenses/by-sa/3.0/deed.en}{CC BY-SA 3.0}], via \href{https://commons.wikimedia.org/wiki/File:Sunrise_over_the_sea.jpg}{Wikimedia Commons}). The dolphin and skull images were produced by Picbreeder~\cite{secretan2008picbreeder} and used in~\cite{gaier2019quality}.}
    \label{fig:top_image_by_fn}
\end{figure*}

\subsection{Functions per cell}

        \begin{table*}[!htp]
        \centering
        \setlength\tabcolsep{1.5pt} 
        \begin{tabular}{| c | c | c | c | c | c | c |}
        \hline
         & Mean &  Mean &  Mean &  Mean & Mean & Mean final
        \\
         & \ total unique & cells & jumps  & total & jump & champion
        \\
        & solutions & in ancestry & in ancestry & replacements & proportion & fitness \\ 
        &(95\% CI)&(95\% CI)&(95\% CI)&(95\% CI)&(95\% CI)&(95\% CI)
        \\
        
        \hline 
            Functions per cell  \\
            \hline
            1 & 12.75 ($\pm 0.39$)          & 6.73 ($\pm 1.15$)          & 24.57 ($\pm 5.48$)  & 261 ($\pm 29$) & 47.43\% ($\pm 1.72\%$)   & 1.10 ($\pm 0.04$)\\ 
            3 & \textbf{40.95} ($\pm 5.52$) & 38.59 ($\pm 6.92$)       & 69.33 ($\pm 17.7$)  & 36,199 ($\pm 4,851$) & 96.36\% ($\pm 0.21\%$)   & \textbf{1.22} ($\pm 0.03$)\\ 
            5 & 30.90 ($\pm 4.01$)          & \textbf{48.67} ($\pm 7.01$)& \textbf{100.85} ($\pm 26.09$)& \textbf{41,283} ($\pm 5,949$) & 97.40\% ($\pm 0.14\%$)   & \textbf{1.19} ($\pm 0.03$) \\
            7 & 22.30 ($\pm 4.03$)          & 40.08 ($\pm 10.14$)& \textbf{84.18} ($\pm 26.54$)  & 28,891 ($\pm 7,523$) & 97.78\% ($\pm 0.13\%$) & \textbf{1.21} ($\pm 0.03$)\\
            9 & 16.10 ($\pm 3.85$)          & 38.85 ($\pm 7.58$)& \textbf{94.2} ($\pm 23.82$)  & 28,092 ($\pm 6,766$) & 98.30\% ($\pm 0.09\%$) & \textbf{1.20} ($\pm 0.03$)\\
            11 & 11.75 ($\pm 4.17$)          & 32.27 ($\pm 7.74$)         & \textbf{101.36} ($\pm 10.15$)  & 22,665 ($\pm 6,621$) & \textbf{98.55\%} ($\pm 0.04\%$) & 1.15 ($\pm 0.45$)\\

        \hline
        Allowed jumps with\\
        5 functions per cell\\
        \hline
        Unlimited & 30.90 ($\pm 4.01$)   &  \textbf{48.67} ($\pm 2.23$) &   \textbf{115.44} ($\pm 11.95$) &  \textbf{25,939} ($\pm 1,782$)   &  \textbf{97.40\%} ($\pm 0.14\%$)  & \textbf{1.19} ($\pm 0.03$)\\ 
        One &       \textbf{100} ($\pm 0$)&  44.51 ($\pm 1.90$) & 43.58 ($\pm 3.40$)  &  12,012 ($\pm 474$)  &  96.31\% ($\pm 0.00\%$)  & \textbf{1.18} ($\pm 0.03$)\\ 
        None &      \textbf{100} ($\pm 0$)&  1 ($\pm 0$) &  0 ($\pm 0$) &  4,441 ($\pm 226$)  &  0.0\% ($\pm 0\%$)  & 1.09 ($\pm 0.03$)
        \\
        \hline
        Hillclimber\\
        \hline
        Single-objectives & - & - & - & - & - & 0.68 ($\pm 0.01$)\\ 
        All-objective    & - & - & - & - & - & \textbf{1.0} ($\pm 0.09$)\\ 
        \hline
        \end{tabular}
        \captionof{table}{Statistics on the movement, diversity, and performance of solutions. The highest value in each column is bold, multiple values in a column are bold when other rows are not significantly different from the highest. 
        The greatest diversity of unique solutions at the end of search occurs for subsets of 3 objective functions/cell, while the greatest number of jumps across cells, unique cells in a run champion's lineage, and total successful offspring replacing existing elites are found in conditions with 5 functions/cell.
        }
          \Description{Statistics about the movement and diversity of solutions.}
          \label{tab:numeric_results_detail}
        \end{table*}

    With 5 functions per cell, when we restrict children to only replacing their parent within the same cell, performance drops significantly from 1.19 to 1.09 ($p<0.001$). Similarly, allowing multiple jumps per offspring was the same (for 1, 7, 9 functions per cell) or better (for 3, 5, and 11 functions per cell; $p \leq 0.02$) than only allowing one.

    The no-jumping condition still performed better than the single-objective baseline for 3, 5, 7, 9 and 11 functions per cell ($p\leq \num{9.41e-3}$) but worse than the baseline for 1 function per cell ($p=\num{3.74e-06}$), demonstrating the importance of including unique combinations of objectives within cells.  

\subsection{Goal-switching}
\label{sec:goal_switching}

    Are there any patterns observable suggesting when goal-switching occurs?  Intuitively we might expect cells with overlapping objective to enable more jumping between them.  With 5 functions per cell, the average number of shared functions between any two cells in our trials was $1.87 / 5$ ($\pm.03$). 
    
     When an elite produced a surviving offspring, the replaced cell shared significantly more functions with the parent cell than the overall average (jumping to one: $1.90 / 5$ $ \pm0.03$, $p<0.01$; jumping to any: $1.92 / 5$ $\pm0.03$, $p<0.001$). As expected, when no jumping was allowed, the average was $5.0 / 5$ ($\pm0$, $p<0.001$) since the children can only replace their parent.  Findings for other number of functions per cell were similar, suggesting that shared objectives do enable increased goal-switching.  

    When considering the complete evolutionary history, 5 functions per cell resulted in more unique cells in the genealogy of the final solutions than any other number of functions per cell (Table~\ref{tab:numeric_results_detail}, $p<0.01$). While 1 function per cell resulted in significantly less unique cells in final solution ancestries ($6.73/100 \pm 1.15$) than all other conditions ($p <0.001$). With 5 functions per cell, the path taken by the ancestors of a final solution included almost half of all the cells in the map on average, suggesting that a wide variety of cells serve as stepping stones en route to a final solution.  Since so many different cells (each with a random subset of objective functions) are present in the evolutionary history of a solution, it is understandable how MOVE can produce solutions that perform well on the mean of all 14 fitness functions, even when not ever being exposed to more than a handful at any given time. 
        
    In addition to more exploration of the map, with 5 functions per cell, there was significantly more total replacements (i.e. successful offspring) over the 1000 generations ($p \leq \num{0.01}$; Table~\ref{tab:numeric_results_detail}), followed by 3 functions per cell which had more than the others ($p \leq \num{8.03e-3}$). With only 1 function per cell, there were significantly less replacements than all other conditions ($p \leq \num{6.3e-08}$). As the number of functions per cell increases, the number of votes that a child needs (i.e. the number of objectives it must simultaneously satisfy) to replace an elite increases, perhaps helping to explain why there are fewer replacements with 11 functions per cell.

    Ablating the ability for offspring to jump to a cell other than its parents' (goal-switch) predictably resulted in fewer replacements and more unique solutions. Similarly, limiting the offspring to only be able to jump to one cell resulted in more unique solutions and fewer jumps than trials with unlimited jumps (Table~\ref{tab:numeric_results_detail}).

\subsection{By function}
    
    MOVE champions outperformed the single-objective hillclimber on 9/14 objective functions. Even when we gave the hillclimber far more compute by increasing the number of offspring per generation per hillclimber to 100 each (1400/gen overall across all hillclimbers), MOVE still performed better on 4 objectives ($p \leq 0.03$), the same on 7, and worse on just 3 ($p \leq 1.13e-3$).

\section{Other target images}
\label{other_targets}
     \begin{figure*}[!htb]
      \centering
      \includegraphics[width=\linewidth]{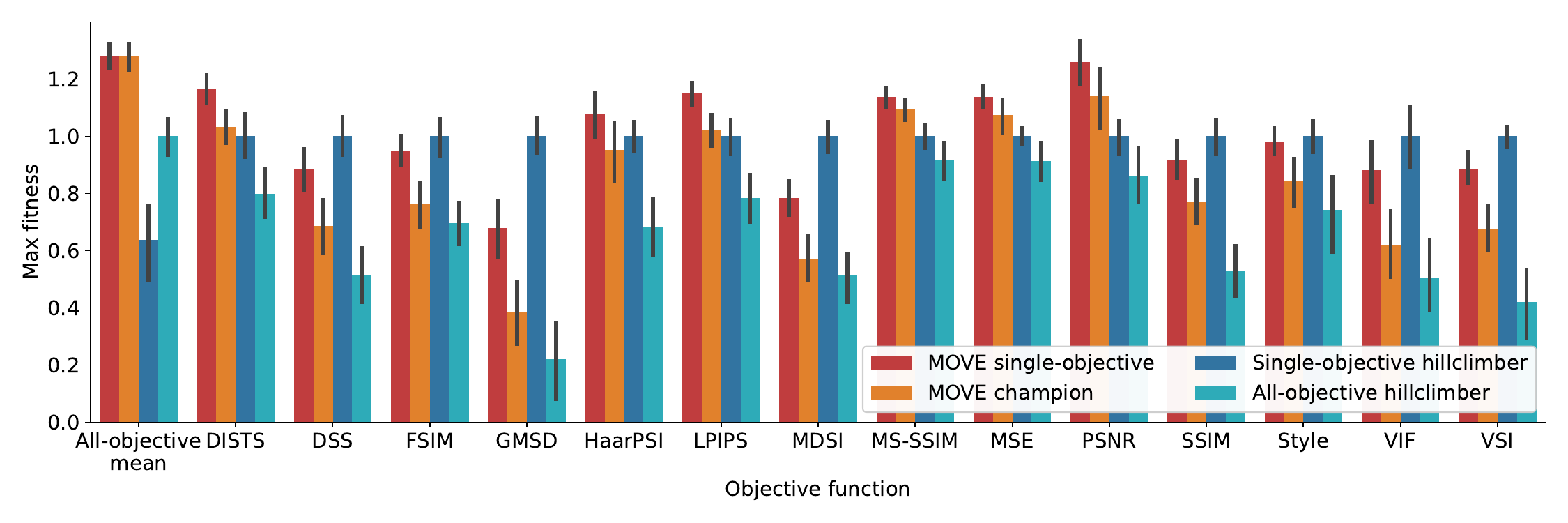}
      \caption{Most fit individuals by objective function, averaged across trials with four different target images, 20 trials per image. Values are transformed so that $1.0$ corresponds with the mean fitness for the corresponding hillclimber. MOVE finds a single champion solution (the highest all-objective mean fitness) from each run that perform the same or better than the solutions found by the all-objective-hillclimber on mean aggregate fitness.  Unlike with the sunrise image on its own, these all-objective MOVE champions only significantly outperform the all-objective-hillclimber solutions on 10 of the 14 individual objectives.  
      Averaged across all images, MOVE all-objective champions also perform as well as, or better, than just 5 of the 14 single-objective hillclimber solutions on the single objective functions for which they were trained. 
      The top performing MOVE solutions on each individual objective (which are trained on the map of all objectives, but may not necessarily be the all-objective champion from that run) score higher or the same as the corresponding single-objective hillclimber on 7 of the 14 objectives (all $p<.001$). 
      The single-objective hillclimber still always significantly outperforms the all-objective hillclimber on single objectives by focusing on one objective. 
      }
      \Description{Max fitness by objective function.}
        \label{fig:max_fits_by_function_all}
    \end{figure*}
    
 The results reported in this work were from trials with the sunrise target image. Hyperparameters were highly tuned to that specific simple image. Trials with more complicated images show that MOVE champions still outperform the all-objective hillclimber, but no longer are very likely to outperform the single-objective hillclimbers (Fig.~\ref{fig:max_fits_by_function_all}). These more complicated images are presumably more difficult to find stepping stones for, and a version of MOVE that is tuned for the sunrise image does not generalize its ability to escape local optima. More research is required to understand how MOVE can be generalized to other target images and other tasks.
 


\section{Limitations and Future work}
\label{sec:future_work}
The specific objective functions in this work was mostly chosen for convenience of implementation in~\cite{kastryulin2022piq}. 
Preliminary exploratory data analysis suggested that many, but not all, pairwise combinations of objectives were well aligned and complementary -- such that, on average, increasing one would also increase the other.  This matches the intuition of the different fitness functions being alternative ways to assess the similarity of two images (the evolved solution to the target images).  Though there are notable differences in how the different Image Quality Assessment metrics make this assessment, and the general intuition that color-based, texture-based, deep-learning-content/encoding-based, etc. metrics being different; the visual inspection of the single-objective hillclimber evolved solutions for each (e.g. Figure~\ref{fig:top_image_by_fn}); and the lower performance of the all-objective hillclimber all similarly suggest that these are not perfectly complementary objectives.  It is not clear how much the alignment between objectives influences the MOVE algorithm, which is a major limitation for the generalizability of this work.  
Future work should consider more carefully crafted fitness metrics which may be more or less compatible with one another, as well as a variety of complex real-world many-objective optimization problems.
Additionally, recombination is an important tool for evolutionary many-objective optimization algorithms and was not implemented for this paper. Future work should consider different encodings, mutation operators and recombination.

\end{document}